\documentclass[runningheads]{asmda2007}

\usepackage{asmda2007References} % For References style

\usepackage{graphicx} % standard LaTeX graphics tool
\usepackage{amssymb}
\usepackage{amsmath}
\newcommand{\ps}[1]{{\langle #1 \rangle}}
\def\nbOne{\mathchoice {\rm 1\mskip-4mu l} {\rm 1\mskip-4mu l}
{\rm 1\mskip-4.5mu l} {\rm 1\mskip-5mu l}}
\begin{document}
%
% \title* lets you specify the title of your manuscript.
% Use \protect\newline to force a line break in your title.
\title*{Scale-sensitive $\Psi$-dimensions: the Capacity
Measures for Classifiers Taking Values in $\mathbb{R}^Q$}
%
% \toctitle specifies the title as will be printed in the table of 
% contents.
% Use \protect\newline to force a line break in your title.
\toctitle{Scale-sensitive $\Psi$-dimensions: the Capacity
Measures for Classifiers Taking Values in $\mathbb{R}^Q$}
%
% \titlerunning defines the title in the running head. Abbreviate
% your title, if the full title is too long to fit in the running 
% head.
\titlerunning{VC dimensions for Classifiers Taking Values in $\mathbb{R}^Q$}
%  
% \authors specifies the authors. Please use initials. Authors are 
% seperated by the \and command. Use the \inst{1} and \inst{2} commands 
% to define the reference mark to your affiliation if needed.
\author{
  Yann Guermeur
}
%
% The following command allow each of the authors to appear 
% in the author index.
% \index{Author, A.}
\index{Guermeur, Y.}
%
% \authorrunning specifies the author name(s) in the running head.
% If there are more than two authors, please abbreviate author list
% (e.g., Skiadas  et al.) for running head.
\authorrunning{Guermeur}
%
% The \institute command lets you specify  your affiliation and
% your address. Seperate two or more different affiliations by the
% \and command. 
\institute{
  LORIA-CNRS\\
  Campus Scientifique, BP~239\\
  54506 Vand\oe uvre-l\`es-Nancy Cedex, France\\
  (e-mail: {\tt Yann.Guermeur@loria.fr})
}

% Typeset the title
\maketitle             

\begin{abstract}
Bounds on the risk play a crucial role
in statistical learning theory. They usually
involve as capacity measure of the model studied the VC
dimension or one of its extensions. In classification,
such ``VC dimensions'' exist for models taking values
in $\left \{0, 1 \right \}$, $\left \{1, \ldots, Q \right \}$
and $\mathbb{R}$. We introduce the generalizations appropriate
for the missing case, the one of models with values in $\mathbb{R}^Q$.
This provides us with a new guaranteed risk for M-SVMs
which appears superior to the existing one.
\end{abstract}

\section{Introduction}

Vapnik's statistical learning theory~\cite{Vap98} deals with three
types of problems: pattern recognition, regression estimation
and density estimation. However,
the theory of bounds has primarily been developed for the computation
of dichotomies only. Central in this theory is the notion of ``capacity''
of classes of functions. In the case of binary classifiers,
the measure of this capacity
is the famous Vapnik-Chervonenkis (VC) dimension. Extensions
have also been proposed for real-valued bi-class models and multi-class models
taking theirs values in the set of categories.
Strangely enough, no generalized VC dimension was available so far for
$Q$-category classifiers taking their values in $\mathbb{R}^Q$.
This was all the more unsatisfactory as many classifiers exhibit
this property, such as the multi-layer perceptrons, or the multi-class
support vector machines (M-SVMs). 
In this paper, the scale-sensitive $\Psi$-dimensions are introduced
to fill this gap.
A generalization of Sauer's lemma \cite{Sau72} is given, 
which relates the covering
numbers appearing in the standard guaranteed risk for
large margin multi-category discriminant models to one of these dimensions,
the margin Natarajan dimension. This latter dimension is then bounded
from above for the architecture shared by all the M-SVMs proposed so far.
This provides us with a sharper bound on their sample complexity.
The organization of the paper is as follows.
Section~\ref{sec:basic_bound} introduces
the basic bound on the risk of
large margin multi-category discriminant models.
In Section~\ref{sec:gamma_psi-dims}, the scale-sensitive $\Psi$-dimensions
are defined, and the generalized Sauer lemma is formulated.
The upper bound on
the margin Natarajan dimension of the M-SVMs is then described
in Section~\ref{sec:M-SVMs}.
For lack of space, proofs are omitted. They can be found in \cite{Gue04}.

\section{Basic theory of large margin $Q$-category classifiers}
\label{sec:basic_bound}

We consider $Q$-category pattern recognition problems,
with $3 \leq Q < \infty$.
A pattern is represented by its description $x \in {\cal X}$
and the set of categories ${\cal Y}$ is identified with the set of indices of 
the categories, $\left \{ 1, \ldots, Q \right \}$. The link between patterns
and categories is supposed to be probabilistic.
${\cal X}$ and ${\cal Y}$ are probability spaces,
and ${\cal X} \times {\cal Y}$ is endowed with a probability measure $P$,
fixed but unknown. Let $\left ( X, Y \right )$ be a random pair
distributed according to $P$.
Training consists in using a $m$-sample 
$s_m = \left ( (X_i, Y_i) \right )_{1 \leq i \leq m}$ of
independent copies of $\left ( X, Y \right )$ to select,
in a given class of functions ${\cal G}$, a function
classifying data in an optimal way. The criterion
to be optimized, the {\em risk}, is the expectation 
with respect to $P$ of a given loss function. The way
the functions in ${\cal G}$ perform classification must be specified.
We consider classes of functions from ${\cal X}$ into $\mathbb{R}^Q$.
$g = \left ( g_k \right )_{1 \leq k \leq Q} \in {\cal G}$
assigns $x \in {\cal X}$ to the category $l$ if and only
if $g_l(x) > \max_{k \neq l} g_k(x)$. Cases of ex \ae quo are treated
as errors. This calls for the choice of a loss function 
$\ell$ defined on
${\cal G} \times {\cal X} \times {\cal Y}$ by
$\ell \left ( y, g(x) \right )  = 
\nbOne_{ \left \{ g_y(x) \leq \max_{k \neq y} g_k(x) \right \}}$.
The risk of $g$ is then given by:
$$
R(g) = \mathbb{E} \left [ \ell \left ( Y, g \left ( X \right ) \right ) \right ]
= \int_{{\cal X} \times {\cal Y}} 
\nbOne_{ \left \{ g_y(x) \leq \max_{k \neq y} g_k(x) \right \}}
dP(x, y).
$$
This study deals with large margin classifiers,
when the underlying notion of multi-class margin is the following one.
\begin{definition}[Multi-class margin]
\label{definition:functional_margin}
Let $g$ be a function from ${\cal X}$ into $\mathbb{R}^Q$. Its {\em margin}
on $(x, y) \in {\cal X} \times {\cal Y}$, ${\cal M}_{xy}(g, x, y)$, is given by:
$$
{\cal M}_{xy}(g, x, y) = \frac{1}{2} 
\left \{ g_{y}(x) - \max_{k \neq y} g_k(x) \right \}.
$$
\end{definition}
Basically, the central elements to assign a pattern to a category
and to derive a level of confidence in this assignation are
the index of the highest output and the difference between this output
and the second highest one. The class of functions
of interest is thus the image of ${\cal G}$ by application
of an appropriate operator.
Two such ``margin operators'' are considered here, $\Delta$ and $\Delta^*$.
\begin{definition}[$\Delta$ operator]
Define $\Delta$ as an operator on ${\cal G}$ such that:
$$
\begin{array}{l l}
\Delta: & {\cal G} \longrightarrow \Delta {\cal G}\\
& g \mapsto \Delta g = \left ( \Delta g_k \right )_{1 \leq k \leq Q}
\end{array}
$$
$$
\forall x \in {\cal X}, \;
\Delta g(x) = 
\frac{1}{2} \left ( g_k(x) - \max_{l \neq k} g_l(x) \right )_{1 \leq k \leq Q}.
$$
\label{definition:first_margin_operator}
\end{definition}
$\forall \left ( g, x \right ) \in {\cal G} \times {\cal X}$,
let ${\cal M}_x(g, x) = \max_k \Delta g_k(x)$.
\begin{definition}[$\Delta^*$ operator]
Define $\Delta^*$ as an operator on ${\cal G}$ such that:
$$
\begin{array}{l l}
\Delta^*: & {\cal G} \longrightarrow \Delta^* {\cal G}\\
& g \mapsto \Delta^* g = \left ( \Delta^* g_k \right )_{1 \leq k \leq Q}
\end{array}
$$
$$
\forall x \in {\cal X}, \;
\Delta^* g(x) =  \left ( \text{sign} \left ( \Delta g_k(x) \right ) \cdot
{\cal M}_x(g, x) \right )_{1 \leq k \leq Q}.
$$
\label{definition:second_margin_operator}
\end{definition}
In the sequel, $\Delta^{\#}$ is used in place of $\Delta$ and $\Delta^*$
in the formulas that hold true for both operators.
The empirical margin risk is defined as follows.
\begin{definition}[Margin risk]
Let $\gamma \in \mathbb{R}_+^*$.
The {risk with margin $\gamma$} of $g$, $R_{\gamma}(g)$,
and its empirical estimate on $s_m$, $R_{\gamma, s_m}(g)$, are defined as:
$$
R_{\gamma}(g)= \int_{{\cal X} \times {\cal Y}} 
\nbOne_{ \left \{ \Delta^{\#} g_y(x) < \gamma \right \}} dP(x,y), \;\;
R_{\gamma, s_m}(g)=
\frac{1}{m} \sum_{i=1}^m \nbOne_{
\left \{ \Delta^{\#} g_{Y_i} \left ( X_i \right ) < \gamma \right \} }.
$$
\label{definition:risque_emp}
\end{definition}
For technical reasons, it is useful to squash
the functions $\Delta^{\#} g_k$ as much as possible without altering the value
of the empirical margin risk.
This is achieved by application of another operator.
\begin{definition}[$\pi_\gamma$ operator \cite{Bar98}]
\label{definition:pi_op}
For $\gamma \in \mathbb{R}_+^*$,
define $\pi_\gamma$ as an operator on ${\cal G}$ such that:
$$
\begin{array}{l l}
\pi_\gamma: & {\cal G} \longrightarrow \pi_\gamma {\cal G}\\
& g \mapsto \pi_\gamma g = \left ( \pi_\gamma g_k \right ))_{1 \leq k \leq Q}
\end{array}
$$
$$
\forall x \in {\cal X}, \;
\pi_\gamma g(x) = \left (
\text{sign} \left ( g_k(x) \right ) \cdot \min \left (|g_k(x)|, \gamma \right )
\right )_{1 \leq k \leq Q}.
$$
\end{definition}
Let $\Delta_{\gamma}^{\#}$ denote $\pi_{\gamma} \circ \Delta^{\#}$ and
$\Delta_{\gamma}^{\#}{\cal G}$ be defined as the set
of functions $\Delta_{\gamma}^{\#}g$. The capacity of 
$\Delta_{\gamma}^{\#}{\cal G}$ is characterized by its covering numbers.
\begin{definition}[$\epsilon$-cover, $\epsilon$-net and covering numbers]
Let $(E,\rho)$ be a pseudo-metric space, $E' \subset E$ and
$\epsilon \in \mathbb{R}_+^*$.
An {\em $\epsilon$-cover} of $E'$ is a coverage of $E'$
with open balls of radius $\epsilon$
the centers of which belong to $E$. These centers form an {\em $\epsilon$-net}
of $E'$. A {\em proper $\epsilon$-net} of $E'$ is an
{\em $\epsilon$-net} of $E'$ included in $E'$.
If $E'$ has an $\epsilon$-net of finite cardinality, then its
{\em covering number}
${\cal N}(\epsilon, E', \rho)$ is the smallest cardinality
of its $\epsilon$-nets. If there is no such finite cover, then
the covering number is defined to be $\infty$.
${\cal N}^{(p)}(\epsilon, E', \rho)$ will designate the covering number
of $E'$ obtained by considering proper $\epsilon$-nets only.
\label{definition:cov_num}
\end{definition}
The covering numbers of interest use the following pseudo-metric:
\begin{definition}[functional pseudo-metric]
Let $\cal G$ be a class of functions from ${\cal X}$ into $\mathbb{R}^Q$.
For a set $s_{{\cal X}^n} \subset {\cal X}$ of cardinality $n$,
define the pseudo-metric
$d_{\ell_\infty, \ell_\infty(s_{{\cal X}^n})}$ on $\cal G$ as:
$$
\forall (g, g')\in {\cal G}^2,\;
d_{\ell_\infty, \ell_\infty(s_{{\cal X}^n})}(g, g') = 
\max_{x \in s_{{\cal X}^n}}
\left \| g(x) - g'(x) \right \|_{\infty}.
$$
\label{definition:distance}
\end{definition}
Let ${\cal N}_{\infty, \infty}^{(p)}(\epsilon,\Delta_{\gamma}^{\#}
{\cal G}, n) =
\sup_{s_{{\cal X}^n} \subset {\cal X}} 
{\cal N}^{(p)}(\epsilon, \Delta_{\gamma}^{\#} {\cal G},
d_{\ell_\infty, \ell_\infty(s_{{\cal X}^n})})$.
The following theorem
extends to the multi-class case Corollary~9 in \cite{Bar98}.
\begin{theorem}[Theorem~1 in \cite{Gue04}]
Let $s_m$ be a $m$-sample of examples independently drawn from a probability
distribution on ${\cal X} \times {\cal Y}$.
With probability at least $1 - \delta$,
for every value of $\gamma$ in $(0, 1]$,
the risk of any function $g$ in a class ${\cal G}$
is bounded from above by:
\begin{equation}
R(g) \leq R_{\gamma, s_m}(g) +
\sqrt{\frac{2}{m}
\left(\ln \left ( 2{\cal N}_{\infty, \infty}^{(p)}
(\gamma/4,\Delta_{\gamma}^{\#} {\cal G}, 2m) \right )
+ \ln \left (\frac{2}{\gamma \delta}\right ) \right)} + \frac{1}{m}.
\label{eq:borne}
\end{equation}
\label{theorem:risque_garanti}
\end{theorem}
Studying the sample complexity of a classifier ${\cal G}$ can thus amount to
computing an upper bound on ${\cal N}_{\infty, \infty}^{(p)}
(\gamma/4,\Delta_{\gamma}^{\#} {\cal G}, 2m)$. In \cite{GueMauSur05},
we reached this goal by relating these numbers to the entropy numbers
of the corresponding evaluation operator. In the present paper, we
follow the traditional path of VC bounds, by making use of a generalized
VC dimension.

\section{Bounding covering numbers in terms of the margin Natarajan dimension}
\label{sec:gamma_psi-dims}

The $\Psi$-dimensions are the generalized VC dimensions that characterize
the learnability of classes of $\left \{1, \ldots, Q \right \}$-valued
functions.
\begin{definition}[$\Psi$-dimensions \cite{BenCesHauLon95}]
Let ${\cal F}$ be a class of functions on a set ${\cal X}$
taking their values in the finite set $\left \{1, \ldots, Q \right \}$.
Let $\Psi$ be a set of mappings $\psi$ from
$\left \{1, \ldots, Q \right \}$ into $\left \{ -1, 1, * \right \}$,
where $*$ is thought of as a null element.
A subset $s_{{\cal X}^n} = \left \{ x_i:  1 \leq i \leq n \right \}$
of ${\cal X}$
is said to be {\em $\Psi$-shattered} by ${\cal F}$ if there is a mapping
$\psi^n = 
\left ( \psi^{(1)}, \ldots, \psi^{(i)}, \ldots, \psi^{(n)} \right )$ 
in $\Psi^n$ such that
for each vector $v_y$ of $\left \{ -1, 1 \right \}^n$, there is
a function $f_y$ in  ${\cal F}$ satisfying
$$
\left ( \psi^{(i)} \circ f_y(x_i) \right )_{1 \leq i \leq n} = v_y.
$$
The {\em $\Psi$-dimension} of ${\cal F}$, denoted by 
$\Psi\mbox{-dim}({\cal F})$, is
the maximal cardinality of a subset
of ${\cal X}$ $\Psi$-shattered by ${\cal F}$, if it is finite, or infinity
otherwise.
\end{definition}
One of these dimensions needs to be singled out, the Natarajan dimension.
\begin{definition}[Natarajan dimension \cite{BenCesHauLon95}]
\label{definition:Natarajan_dimension}
Let ${\cal F}$ be a class of functions on a set ${\cal X}$
taking their values in $\left \{ 1, \ldots, Q \right \}$.
The {\em Natarajan dimension} of ${\cal F}$, $\mbox{N-dim}({\cal F})$, 
is the $\Psi$-dimension of ${\cal F}$
in the specific case where $\Psi$ is the set of $ Q (Q-1) $ mappings
$\psi_{k,l}$, $(1 \leq k \neq l \leq Q)$, such that $\psi_{k,l}$
takes the value $1$ if its argument is equal to $k$, the value  $-1$
if its argument is equal to $l$, and $*$
otherwise.
\end{definition}
The fat-shattering dimension
characterizes the uniform Glivenko-Cantelli classes
among the classes of real-valued functions.
\begin{definition}[fat-shattering dimension \cite{AloBenCesHau97}]
Let ${\cal G}$ be a class of functions from ${\cal X}$ into $\mathbb{R}$.
For $\gamma \in \mathbb{R}_+^*$, 
$s_{{\cal X}^n} = \left \{ x_i: 1 \leq i \leq n \right \} \subset {\cal X}$
is said to be {\em ${\gamma}$-shattered} by ${\cal G}$ if
there is a vector $v_b =  \left ( b_i \right ) \in \mathbb{R}^n$ such that,
for each vector $v_y = \left ( y_i \right ) \in
\left \{ -1, 1 \right \}^n$, there is a function
$g_y \in {\cal G}$ satisfying
$$
\forall i \in \left \{1, \ldots, n \right \}, \;
y_i \left( g_{y}(x_i) - b_i \right) \geq \gamma.
$$
The {\em fat-shattering dimension} of
${\cal G}$, $P_{\gamma}\mbox{-dim}\left({\cal G}\right)$,
is the maximal cardinality of a subset of ${\cal X}$
${\gamma}$-shattered by ${\cal G}$, if it is finite, or infinity
otherwise.
\label{definition:fat_shattering_dim}
\end{definition}
Given the results available for the $\Psi$-dimensions and the
fat-shattering dimension, it appears natural, 
to study the generalization capabilities of classifiers taking values in
$\mathbb{R}^Q$, to consider the use of capacity measures
obtained as mixtures of the two concepts, namely
scale-sensitive $\Psi$-dimensions.
\begin{definition}[$\Psi$-dimension with margin $\gamma$]
\label{definition:guermeur_dim}
Let ${\cal G}$ be a class of functions on a set ${\cal X}$
taking their values in $\mathbb{R}^Q$.
Let $\Psi$ be a family of mappings $\psi$ from
$\left \{1, \ldots, Q \right \}$
into $\left \{ -1, 1, * \right \}$. For $\gamma \in \mathbb{R}_+^*$,
a subset $s_{{\cal X}^n} = \left \{ x_i: 1 \leq i \leq n \right \}$
of ${\cal X}$
is said to be {\em $\gamma$-$\Psi$-shattered} 
by $\Delta^{\#} {\cal G}$ if there is a mapping
$\psi^n = 
\left ( \psi^{(1)}, \ldots, \psi^{(i)}, \ldots, \psi^{(n)} \right )$ 
in $\Psi^n$ and a vector $v_{b} =  \left ( b_i \right )$
in $\mathbb{R}^n$
such that, for each vector
$v_y = \left ( y_i \right )$ of $\left \{ -1, 1 \right \}^n$, 
there is a function $g_y$ in  ${\cal G}$ satisfying
$$
\forall i \in \left \{1, \ldots, n \right \}, \;
\left\{\begin{array}{lrlrl}
\text{if} \; y_i = & 1, & \exists k:
\psi^{(i)}(k)  =  & 1  & \wedge \;\;
\Delta^{\#} g_{y, k}(x_i) - b_{i} \geq \gamma \\
\text{if} \; y_i = & -1, & \exists l:
\psi^{(i)}(l)  =  & -1 & \wedge \;\;
\Delta^{\#} g_{y, l}(x_i) + b_{i} \geq \gamma \\
\end{array}
\right..
$$
The {\em $\gamma$-$\Psi$-dimension} of $\Delta^{\#} {\cal G}$,
$\Psi$-dim$(\Delta^{\#} {\cal G}, \gamma)$, 
is the maximal cardinality of a subset of
${\cal X}$ $\gamma$-$\Psi$-shattered by $\Delta^{\#} {\cal G}$, if it is finite,
or infinity otherwise.
\end{definition}
The margin Natarajan dimension is defined accordingly.
\begin{definition}[Natarajan dimension with margin $\gamma$]
\label{definition:guermeur_Nat_dim}
Let ${\cal G}$ be a class of functions on a set ${\cal X}$
taking their values in $\mathbb{R}^Q$. For $\gamma \in \mathbb{R}_+^*$,
a subset $s_{{\cal X}^n} = \left \{ x_i: 1 \leq i \leq n \right \}$
of ${\cal X}$
is said to be {\em $\gamma$-N-shattered}
by $\Delta^{\#} {\cal G}$ if there is a set
$I(s_{{\cal X}^n}) =
\left  \{ \left (i_1(x_i), i_2(x_i) \right ): 1 \leq i \leq n \right \}$
of $n$ pairs of distinct indices in $\left \{1, \ldots, Q \right \}$
and a vector
$v_{b} =  \left ( b_i \right )$ in $\mathbb{R}^n$
such that,
for each binary vector $v_y = \left ( y_i \right ) \in
\left \{ -1, 1 \right \}^n$, there is a function
$g_y$ in ${\cal G}$ satisfying
$$
\forall i \in \left \{1, \ldots, n \right \}, \;
\left\{\begin{array}{lrl}
\text{if} \; y_i = & 1, &
\Delta^{\#} g_{y, i_1(x_i)}(x_i) - b_{i} \geq \gamma \\
\text{if} \; y_i = & -1, &
\Delta^{\#} g_{y, i_2(x_i)}(x_i) + b_{i} \geq \gamma \\
\end{array}
\right..
$$
The {\em Natarajan dimension with margin $\gamma$} of the class
$\Delta^{\#} {\cal G}$, 
$\mbox{N-dim}(\Delta^{\#} {\cal G}, \gamma)$, is the maximal
cardinality of a subset of ${\cal X}$ $\gamma$-N-shattered by
$\Delta^{\#} {\cal G}$, if it is finite, or infinity otherwise.
\end{definition}
For this scale-sensitive $\Psi$-dimension, the connection with
the covering numbers of interest, or generalized Sauer lemma,
is the following one.
\begin{theorem}[Theorem~4 in \cite{Gue04}]
\label{theorem:covering_margin_Natarajan_final}
Let ${\cal G}$ be a class of functions from a domain ${\cal X}$ into
$\mathbb{R}^Q$. For every value of $\gamma$ in $(0, 1]$ and every
$m \in \mathbb{N}^*$ satisfying
$2m \geq \mbox{N-dim}\left (\Delta_{\gamma} {\cal G}, \gamma / 24 \right )$,
the following bound is true:
\begin{equation}
\label{eq:covering_margin_Natarajan_final}
{\cal N}_{\infty, \infty}^{(p)}(\gamma/4,\Delta_{\gamma}^* {\cal G}, 2m) <
2 \left ( 288 \; m \; Q^2 (Q-1) \right )^{
\lceil d \log_2 \left ( 23 e m Q (Q-1) / d  \right ) \rceil }
\end{equation}
where $d = \mbox{N-dim}\left (\Delta_{\gamma} {\cal G}, \gamma / 24 \right )$.
\end{theorem}
This theorem is the central result of the paper
(and the novelty in the revised version of \cite{Gue04}).
What makes it a nontrivial $Q$-class extension of Lemma~3.5
in \cite{AloBenCesHau97} is the presence of both margin
operators. The reason why $\Delta^*$ appears
in the covering number instead of $\Delta$ is the very principle at the
basis of all the variants of Sauer's lemma:
two functions separated with respect to the functional pseudo-metric used
(here $d_{\ell_\infty, \ell_\infty(s_{{\cal X}^n})}$)
shatter (at least) one point in $s_{{\cal X}^n}$. 
This is true for $\Delta_{\gamma}^* {\cal G}$,
or more precisely its $\eta$-discretization, 
not for $\Delta_{\gamma} {\cal G}$
(see Section~5.3 in \cite{Gue04} for details).
One can derive a variant of
Theorem~\ref{theorem:covering_margin_Natarajan_final} involving
$\mbox{N-dim}\left (\Delta_{\gamma}^* {\cal G}, \gamma / 24 \right )$.
This alternative is however of lesser interest, for reasons
that will appear below.

\section{Margin Natarajan dimension of the M-SVMs}
\label{sec:M-SVMs}
We now compute an upper bound on the margin Natarajan dimension of interest
when ${\cal G}$ is the class of functions computed by the M-SVMs.
These large margin classifiers are built around a Mercer kernel.
Let $\kappa$ be such a kernel on ${\cal X}$
and $\left ( H_{\kappa}, \ps{.,.}_{H_{\kappa}} \right )$ the corresponding 
reproducing kernel Hilbert space (RKHS) \cite{Aro50}.
Let $\Phi$ be any of the mappings on ${\cal X}$ satisfying:
\begin{equation}
\forall \left (x, x' \right ) \in {\cal X}^2, \;
\kappa \left ( x, x' \right ) =  \ps{\Phi \left ( x \right ), 
\Phi \left ( x' \right )},
\label{eq:kernel_trick}
\end{equation}
where $ \ps{.,.}$ is the dot product of the $\ell_2$ space.
``The'' feature space traditionally designates
any of the Hilbert spaces 
$\left ( E_{\Phi \left ( {\cal X} \right )}, \ps{.,.} \right )$
spanned by the $\Phi \left ( {\cal X} \right )$. By definition of a RKHS,
${\cal H} = \left ( \left ( H_{\kappa}, \ps{.,.}_{H_{\kappa}} \right ) 
+ \left \{ 1 \right \} \right )^Q$
is the class of functions
$h = \left ( h_k \right )_{1 \leq k \leq Q}$ from ${\cal X}$ into
$\mathbb{R}^Q$ of the form:
$$
h(.) = 
\left ( \sum_{i=1}^{l_k} \beta_{ik} \kappa \left ( x_{ik}, . \right ) + b_k 
\right )_{1 \leq k \leq Q}
$$
where the $x_{ik}$ are elements of ${\cal X}$ (the $\beta_{ik}$ and $b_k$
are scalars), as well as the limits of these functions when the sets
$\left \{ x_{ik}: 1 \leq i \leq l_k \right \}$ become dense in ${\cal X}$
in the norm induced by the dot product. Due to
(\ref{eq:kernel_trick}), ${\cal H}$ can also be seen as a multivariate
affine model on $\Phi \left ( {\cal X} \right )$. Functions $h$ can
then be rewritten as:
$$
h(.) = \left ( \ps{w_k, .} + b_k \right )_{1 \leq k \leq Q}
$$
where vectors $w_k$ are elements of $E_{\Phi \left ( {\cal X} \right )}$.
They are thus described by the pair 
$\left (\mathbf{w}, \mathbf{b} \right )$ with
$ \mathbf{w} = \left ( w_k \right )_{1 \leq k \leq Q}$
and $\mathbf{b} = \left ( b_k \right )_{1 \leq k \leq Q}$.
Let $\bar{\cal H}$ stand for the product space $H_{\kappa}^Q$.
Its norm $\|.\|_{\bar{\cal H}}$ is given by 
$\left \| \bar{h} \right \|_{\bar{\cal H}} =
\sqrt{\sum_{k=1}^Q \|w_k \|^2} = \left \| \mathbf{w} \right \|$.
\begin{definition}[M-SVM]
A M-SVM is a large margin multi-category discriminant model
obtained by minimizing over the hyperplane $\sum_{k=1}^Q h_k = 0$ of ${\cal H}$
an objective function of the form:
$$
J \left ( h \right ) = \sum_{i=1}^m \ell_{\text{M-SVM}} \left ( y_i,  
h \left ( x_i \right ) \right ) +
\lambda \left \| \mathbf{w} \right \|^2
\label{eq:obj_general}
$$
where the empirical term, used in place of the empirical risk,
involves a loss function $\ell_{\text{M-SVM}}$ which is convex.
\end{definition}
The M-SVMs only differ in the nature of $\ell_{\text{M-SVM}}$.
The specification of this function is such that the
introduction of the penalizer $\left \| \mathbf{w} \right \|^2$
tends to maximize a notion of margin
directly connected with the one of
Definition~\ref{definition:functional_margin}.
The formulation of the generalized Sauer lemma provided here
(Theorem~\ref{theorem:covering_margin_Natarajan_final}) is the one
obtained under the weakest hypotheses.
Proceeding as in the bi-class case, we express below a bound
on the margin Natarajan dimension of the M-SVMs as a function of
the volume occupied by data in $E_{\Phi \left ( {\cal X} \right )}$
 and constraints on 
$\left (\mathbf{w}, \mathbf{b} \right )$, thus restricting 
the study to functions with a well-defined range.
In that case, a variant of 
Theorem~\ref{theorem:covering_margin_Natarajan_final} can be derived
from Lemma~7 in \cite{Gue04} which does not involve $\pi_\gamma$ but
relates the covering numbers of $\Delta^* {\cal G}$ to
the margin Natarajan dimension of $\Delta {\cal G}$. Its use
for M-SVMs is advantageous since
$\mbox{N-dim}\left ( \Delta \bar{\cal H}, \epsilon \right )$
is easier to bound than
$\mbox{N-dim}\left ( \Delta_{\gamma} {\cal H}, \epsilon \right )$
(nonlinearity is difficult to handle).
This change of generalized Sauer lemma calls for
the use of an intermediate formula relating the covering numbers
of $\Delta_{\gamma}^* {\cal H}$ and $\Delta^* \bar{\cal H}$.
It is provided by the following lemma.
\begin{lemma}[Lemmas~9 and 10 in \cite{Gue04}]
Let ${\cal H}$ be the class of functions that a $Q$-category
M-SVM can implement under the hypothesis
$\mathbf{b} \in \left [ -\beta, \beta \right ]^Q$. 
Let $\left ( \gamma, \epsilon \right ) \in \mathbb{R}^2$
satisfy $0 < \epsilon \leq \gamma \leq 1$. Then
\begin{equation}
{\cal N}_{\infty, \infty}^{(p)}(\epsilon, \Delta_{\gamma}^* {\cal H}, m) \leq
\left ( 2 \left \lceil \frac{\beta}{\epsilon} \right \rceil + 1 \right )^Q
{\cal N}_{\infty, \infty}^{(p)}(\epsilon/2, \Delta^* \bar{{\cal H}}, m).
\end{equation}
\end{lemma}
A final theorem then completes the
construction of the guaranteed risk.
\begin{theorem}[Theorem~5 in \cite{Gue04}]
\label{theorem:VC_YG}
Let $\bar{\cal H}$ be the class of functions that a $Q$-category M-SVM
can implement under the hypothesis that $\Phi ( {\cal X} )$
is included in the closed ball of radius
$\Lambda_{\Phi \left ({\cal X} \right )}$ about the origin in
$E_{\Phi \left ({\cal X} \right )}$
and the constraints $1/2 \max_{1 \leq k < l \leq Q} 
\|w_k - w_l\| \leq \Lambda_w$ and $\mathbf{b} = 0$.
Then, for any
positive real value $\epsilon$, the following bound holds true:
\begin{equation}
\label{eq:VC_YG} 
\mbox{N-dim}\left ( \Delta \bar{\cal H}, \epsilon \right )
\leq C_Q^2
\left ( \frac{\Lambda_w \Lambda_{\Phi({\cal X})}}{\epsilon} \right )^2.
\end{equation}
\end{theorem}
The proof follows the line of argument of the
corresponding bi-class result, Theorem~4.6 in \cite{BarSha99}.
This involves a generalization of
Lemma~4.2 which can only be performed for the $\Delta$ operator.
The discussion on the presence of both $\Delta$ and $\Delta^*$ in
Theorem~\ref{theorem:covering_margin_Natarajan_final} is thus completed.
Putting things together, the control term of 
the guaranteed risk decreases with the size
of the training sample as
$\ln(m) \cdot m^{-1/2}$. This represents an improvement over the rate
obtained in \cite{GueMauSur05}, $m^{-1/4}$.

\section{Conclusions and future work}

A new class of generalized VC dimensions dedicated to large margin
multi-category discriminant models has been introduced. They
can be seen either as multivariate extensions of the fat-shattering dimension
or scale-sensitive $\Psi$-dimensions.
Their finiteness (for all positive values of the scale parameter $\gamma$)
is also a necessary and sufficient condition for learnability.
A generalized Sauer lemma
has been provided for one of these capacity measures, the margin Natarajan
dimension. This latter dimension has been bounded from above in the case
where the classifier is a multi-class SVM.
This study provides us with new arguments to support the thesis that
the theory of multi-category pattern recognition cannot be developed
by extending in a straightforward way bi-class results. We are currently
making use of the specificities identified here to extend new concentration
inequalities to the multi-class case with the goal to
obtain improved convergence rates.

\bibliographystyle{asmda2007References}
\bibliography{bibliography-asmda2007}

\begin{thebibliography}{}

\bibitem[\protect\citeauthoryear{Alon \bgroup \em et al.\egroup
  }{1997}]{AloBenCesHau97}
N.~Alon, S.~Ben-David, N.~Cesa-Bianchi, and D.~Haussler.
\newblock Scale-sensitive dimensions, uniform convergence, and learnability.
\newblock {\em Journal of the ACM}, 44(4):615--631, 1997.

\bibitem[\protect\citeauthoryear{Aronszajn}{1950}]{Aro50}
N.~Aronszajn.
\newblock Theory of reproducing kernels.
\newblock {\em {Transactions of the American Mathematical Society}},
  68(3):337--404, 1950.

\bibitem[\protect\citeauthoryear{Bartlett and Shawe-Taylor}{1999}]{BarSha99}
P.L. Bartlett and J.~Shawe-Taylor.
\newblock Generalization performance of support vector machines and other
  pattern classifiers.
\newblock In B.~Sch\"olkopf, C.J.C. Burges, and A.~Smola, editors, {\em
  Advances in Kernel Methods, Support Vector Learning}, pages 43--54. The MIT
  Press, Cambridge, 1999.

\bibitem[\protect\citeauthoryear{Bartlett}{1998}]{Bar98}
P.L. Bartlett.
\newblock The sample complexity of pattern classification with neural networks:
  the size of the weights is more important than the size of the network.
\newblock {\em IEEE Transactions on Information Theory}, 44(2):525--536, 1998.

\bibitem[\protect\citeauthoryear{Ben-David \bgroup \em et al.\egroup
  }{1995}]{BenCesHauLon95}
S.~Ben-David, N.~Cesa-Bianchi, D.~Haussler, and P.M. Long.
\newblock Characterizations of learnability for classes of $\left \{ 0,\ldots,n
  \right \}$-valued functions.
\newblock {\em Journal of Computer and System Sciences}, 50:74--86, 1995.

\bibitem[\protect\citeauthoryear{Guermeur \bgroup \em et al.\egroup
  }{2005}]{GueMauSur05}
Y.~Guermeur, M.~Maumy, and F.~Sur.
\newblock {Model selection for multi-class SVMs}.
\newblock In {\em ASMDA'05}, pages 507--516, 2005.

\bibitem[\protect\citeauthoryear{Guermeur}{2004}]{Gue04}
Y.~Guermeur.
\newblock {Large margin multi-category discriminant models and scale-sensitive
  $\Psi$-dimensions}.
\newblock Technical Report RR-5314, INRIA, http://hal.inria.fr/inria-00070686,
  2004.
\newblock (revised in 2006).

\bibitem[\protect\citeauthoryear{Sauer}{1972}]{Sau72}
N.~Sauer.
\newblock On the density of families of sets.
\newblock {\em Journal of Combinatorial Theory (A)}, 13:145--147, 1972.

\bibitem[\protect\citeauthoryear{Vapnik}{1998}]{Vap98}
V.N. Vapnik.
\newblock {\em Statistical learning theory.}
\newblock John Wiley \& Sons, Inc., N.Y., 1998.

\end{thebibliography}

\end{document}